\pdfoutput=1
\documentclass[conference]{IEEEtran}
\IEEEoverridecommandlockouts
\usepackage{cite}
\usepackage{amsmath,amssymb,amsfonts}
\usepackage{algorithmic}
\usepackage{graphicx}
\usepackage{textcomp}
\usepackage{xcolor}
\usepackage{bibunits}
\usepackage{romannum}
\usepackage{enumerate}
\usepackage{multirow}
\usepackage[
  font = small,
  labelfont = bf,
  tableposition = top
]{caption}
\usepackage{float}
\usepackage{epstopdf}
\usepackage{gensymb}

\usepackage{comment}

\def\BibTeX{{\rm B\kern-.05em{\sc i\kern-.025em b}\kern-.08em
    T\kern-.1667em\lower.7ex\hbox{E}\kern-.125emX}}
\begin{document}

\title{Reconstructing Depth Images of Moving Objects from Wi-Fi CSI Data
}


\author{\IEEEauthorblockN{Guanyu Cao}
\IEEEauthorblockA{\textit{Graduate School of Information Science and Technology} \\
\textit{Osaka University}\\
Osaka, Japan \\
cao.guanyu@ist.osaka-u.ac.jp}
\\
\IEEEauthorblockN{Kazuya Ohara}
\IEEEauthorblockA{\textit{NTT Communication Science Laboratories} \\
\textit{NTTCSL}\\
Kyoto, Japan \\
kazuya.ohara@ntt.com}

\and
\IEEEauthorblockN{Takuya Maekawa}
\IEEEauthorblockA{
\textit{Graduate School of Information Science and Technology} \\
\textit{Osaka University}\\
Osaka, Japan \\
maekawa@ist.osaka-u.ac.jp}
\\
\IEEEauthorblockN{Yasue Kishino}
\IEEEauthorblockA{\textit{NTT Communication Science Laboratories} \\
\textit{NTTCSL}\\
Kyoto, Japan \\
yasue.kishino@ntt.com}
}

\maketitle

\begin{abstract}
This study proposes a new deep learning method for reconstructing depth images of moving objects within a specific area using Wi-Fi channel state information (CSI). The Wi-Fi-based depth imaging technique has novel applications in domains such as security and elder care. However, reconstructing depth images from CSI is challenging because learning the mapping function between CSI and depth images, both of which are high-dimensional data, is particularly difficult.
To address the challenge, we propose a new approach called Wi-Depth.
The main idea behind the design of Wi-Depth is that a depth image of a moving object can be decomposed into three core components: the shape, depth, and position of the target. 
Therefore, in the depth-image reconstruction task, Wi-Depth simultaneously estimates the three core pieces of information as auxiliary tasks in our proposed VAE-based teacher-student architecture, enabling it to output images with the consistency of a correct shape, depth, and position. 
In addition, the design of Wi-Depth is based on our idea that this decomposition efficiently takes advantage of the fact that shape, depth, and position relate to primitive information inferred from CSI such as angle-of-arrival, time-of-flight, and Doppler frequency shift. 
\end{abstract}


\section{Introduction}
\label{intro}



\subsubsection{Background }

In the pervasive computing community, Wi-Fi sensing techniques have attracted much research interest as a passive, non-intrusive, and privacy-preserving sensing approach because Wi-Fi access points (APs) are pervasive in our daily lives, enabling us to easily access Wi-Fi signal information that reflects our daily life contexts from the APs through commercial off-the-shelf (COTS) devices such as smartphones and laptop computers \cite{advantages}. 
In particular, context recognition using Wi-Fi channel state information (CSI) has been actively studied to predict a variety of real-world contexts. 
As primitive contexts, angle-of-arrival (AoA), time-of-flight (ToF), and Doppler frequency shift (DFS) have been estimated based on multi-antenna and multi-subcarrier information in CSI \cite{context}.
For high-level context estimation, CSI has enabled sensing tasks such as positioning, human skeleton estimation, and activity recognition \cite{scenario}.

Recently, as high-dimensional context recognition from Wi-Fi CSI, Wi-Fi imaging has been attracting attention.
Wi-Fi imaging is a device-free, non-vision imaging technique that estimates images via Wi-Fi.
For example, researchers have attempted to estimate the walking footage of a human \cite{wi2vi} and a human appearing at a specific location \cite{csi2image}.
Wi-Fi imaging has novel but essential applications in various domains such as security and elder care.
As for security, such a system can be installed at the entrance of rooms or corridors as surveillance without invasion of privacy.
As for elder care, it could monitor elderly people living alone, and report emergencies such as falling to the ground.
However, as dominant information in RGB or grey-scale images, the textures of objects hardly affect Wi-Fi transmission. 
In addition, the exterior of RGB or grey-scale images is easily changed by light sources and shadows, while Wi-Fi is free from external illumination.
This makes it impossible to reconstruct RGB or grey-scale images from Wi-Fi.


\begin{figure}[t]
 \begin{center}
\includegraphics[scale=.58]{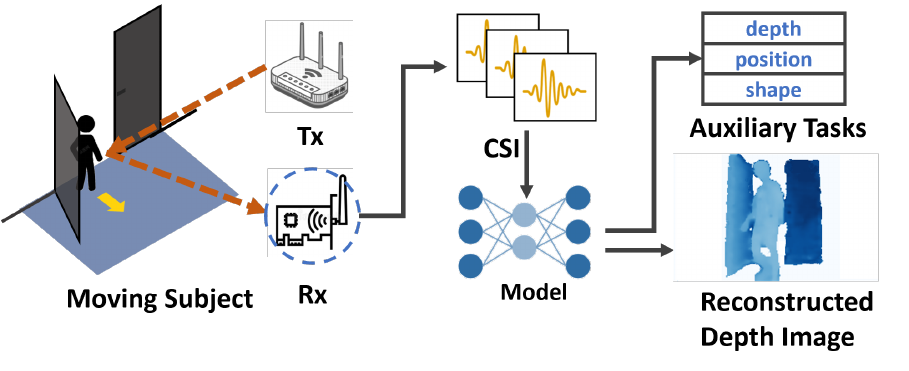}
\caption{Illustration of the proposed approach}
\label{fig:intro}
 \end{center}
\end{figure}

\subsubsection{Goal}
In this study, as a new target of Wi-Fi imaging, we focus on depth images, which measure the distance of objects from the camera.
Specifically, we focus on reconstructing depth images of moving objects without any static components such as backgrounds.
Figure \ref{fig:intro} shows the overall illustration of the proposed approach.
Even though the static components are removed, the above-mentioned applications are still viable because the images of the target of interest are more vital than background components.
We believe that depth reconstruction is more suitable than RGB/grey-scale for Wi-Fi imaging.
Depth images contain 3D physical information including the shape, position, and depth of objects.
Such attributes also affect Wi-Fi transmission by changing reflection, scattering, and occlusion of transmission paths.
As a result, the depth image of an object is associated with a Wi-Fi transmission state, which is represented by CSI.

\subsubsection{Challenges}


However, reconstructing depth images from CSI is still a challenging task.
The main reason is that CSI and depth images are both high-dimensional data.
On the one hand, Wi-Fi has supported Orthogonal Frequency Division Multiplexing (OFDM) since the 802.11a/b standard.
OFDM allows multiple transmitting antennas, receiving antennas, and subcarriers.
As a result, the CSI matrix for a series of packets is of $(N_{Tx} \times N_{Rx} \times N_{subcarrier} \times N_{packet})$ dimensions.
On the other hand, depth images are also high-dimensional, i.e., (width $\times$ height).

Unlike prior Wi-Fi-based context recognition methods such as activity recognition and indoor positioning \cite{widar2, widar3, indotrack, falldefi, gaitway} that output class labels or states of a target from a preset source of possible values, depth images are distributed in a much higher-dimensional space.
Consequently, it is a very tough task to learn the mapping function between CSI and depth images.
Many prior Wi-Fi imaging researches adopted CNN-based autoencoders \cite{wi2vi, csi2image} to convert CSI into an image.
However, the simple autoencoder architecture cannot well grasp three core pieces of information of a depth image, which are the shape, depth, and position of the target as shown in the right image in Figure \ref{fig:intro}, and often fails to estimate consistent images.
For example, the depth and position are correct yet with a blurry shape, or a correct shape at a wrong position.
This is because the latent space of the autoencoder is neither regularized nor efficiently learns core information. 

\subsubsection{Approach}


To address the aforementioned challenges, we propose a new approach called Wi-Depth, which efficiently learns the mapping function from high-dimensional CSI to depth images.
The main idea behind the design of Wi-Depth is that a depth image of a moving object can be decomposed into three core components: the shape, depth, and position of the target. 
Therefore, in the depth-image reconstruction task, Wi-Depth simultaneously estimates the three core pieces of information as auxiliary tasks, enabling it to output images with the consistency of a correct shape, depth, and position. 
In addition, the design of Wi-Depth is based on our idea that this decomposition efficiently takes advantage of the fact that shape, depth, and position relate to primitive information inferred from CSI. For example, as many wireless tracking works have exploited \cite{spotfi, indotrack, widar2}, depth is related to the transmission path length and thus related to ToF. Position is related to coarse AoA and ToF, and shape is related to fine-grained AoA and DFS.

To facilitate the decomposition and reconstruction, Wi-Depth is designed to have a variational autoencoder (VAE)-based teacher-student architecture. The teacher model, which is a VAE, compresses a high-dimensional depth image into a low-dimensional representation and then reconstructs the original depth image. Here, we train the teacher model so that the low-dimensional representation preserves the three core pieces of information, i.e., shape, depth, and position. The decomposition and reconstruction tasks, i.e., depth image to the three core pieces of information and vice versa, are easier than the task of reconstructing a depth image directly from CSI, which is adopted in prior Wi-Fi imaging studies based on non-teacher-student autoencoder architectures. Therefore, accurate latent representations are expected to be learned in the teacher model of Wi-Depth. This facilitates training of the student model, which encodes CSI data into latent representations, because we can leverage reliable latent representations as a target of encoding, unlike the existing autoencoder architecture.

\subsubsection{Contributions}
The contributions of this work are three-fold:
(i) As far as we have investigated, this is the first work to estimate depth images of moving objects from Wi-Fi CSI.
(ii) An efficient architecture is proposed that estimates consistent depth images by decomposing the shape, depth, and position estimates as auxiliary tasks via a VAE-based teacher-student network.
(iii) The effectiveness of the proposed model was validated in four real environments.

\begin{figure*}[ht]
 \begin{center}
\hspace{-0.4cm}\includegraphics[scale=.52]{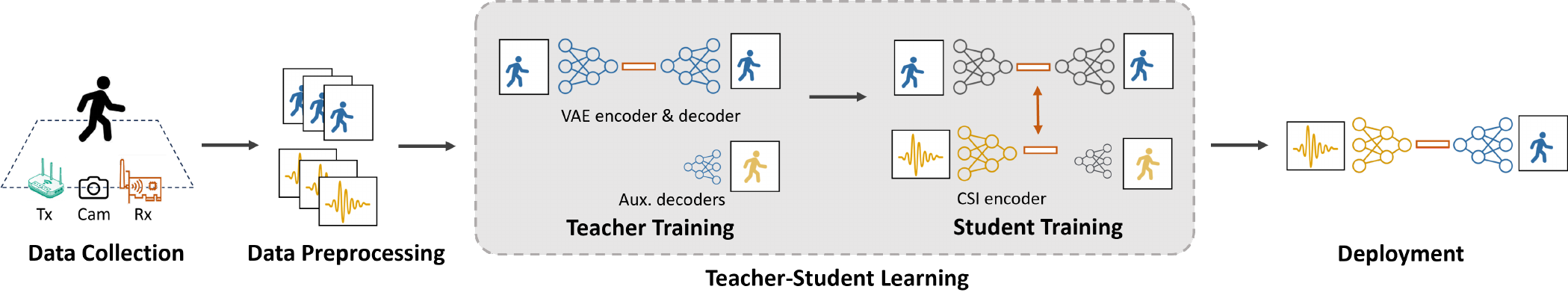}
\caption{Overview of Wi-Depth}
\label{fig:overview}
 \end{center}
\end{figure*}

\section{Related work}
Up till now, the typical approach to Wi-Fi sensing can be categorized into three-fold: Tracking, Detection and Recognition, and Estimation \cite{survey}.
Our proposed method is inspired by related research in all these three aspects, although it concentrates on imaging.

\subsection{Wi-Fi Tracking}
As for tracking methods, CSI could provide the characteristics of the reflected path caused by the subject under observation.
By accessing phase information of the PHY layer, physical features of transmission links can be determined, such as AoA, ToF, and DFS.
The subject can therefore be localized via simple geometric equations \cite{widar2, spotfi, indotrack}.
The phase information of CSI is the major advantage over Received Signal Strength Indicator (RSSI)-based methods.


\subsection{Wi-Fi Detection and Recognition}


Further excavated phase information enables finer-grained human motion detection and recognition approaches.
\cite{widar3, wisee} realize gesture recognition by analyzing the Body Velocity Profile (BVP) extracted from CSI.
\cite{csinet, falldefi} achieve fall detection using deep learning and machine learning methods.
\cite{noninvasive} detects human respiration, allowing it to detect both moving and static humans.
\cite{gaitway, gaitid} achieve gait recognition via auto-correlation and BVP, respectively.

These works significantly enhance the range of Wi-Fi sensing capabilities. 
However, they are constrained to a fixed set of outputs, whether categories or binary flags. 
Although this scheme is well-suited for some real-world applications, we believe that Wi-Fi sensing based on CSI can be explored even further.

\subsection{Estimation by Wi-Fi}
Estimating human pose using Wi-Fi is an eye-catching topic.
\cite{person} estimates the coordinates of key joints of a human.
\cite{densepose} estimates both keypoints and the 2D projection of the 3D human surface, as is done by DensePose \cite{densepose_img}.
\cite{towards3d, winect} estimate the motion update factor that updates the skeleton of a human.
\cite{wimesh} explores human pose reconstruction by estimating the human mesh.
These methods are able to estimate more poses than fixed-target recognition methods.
However, these works only focus on human poses.
The interaction between humans and surrounding objects is not taken into consideration.

\subsection{Wi-Fi Imaging}




A majority of existing Wi-Fi imaging methods start by mimicking traditional radar.
These methods are based on scanning, which is realized by moving the emitter / receiver or beam-forming the scanner medium.
\cite{throughwall} achieves through-wall Wi-Fi imaging by radio frequency (RF) signals projected onto two spatial dimensions, while \cite{aerial} achieves the same goal by binding the transmitter and receiver on two drones and scanning in space.
Works such as \cite{imagingwithwifi, wivi, ris, mobileradar, feasibility} employ antenna arrays or matrices to function as Reconfigurable Intelligent Surface (RIS) or Synthetic Aperture Radar (SAR).
Although these works adhere to the design pipeline of radars, the high cost of antenna arrays or movable bases goes against the daily usage of COTS Wi-Fi.
Moreover, all of these radar-like methods only use the amplitude, or RSSI, of Wi-Fi signals.

As discovered in \cite{fromrssitocsi}, CSI is superior to RSSI in the granularity of transmission path information, which relieves the transceiver from scanning in space.
\cite{csi2image} attempts to reconstruct the authentic scene by directly generating images of humans and surrounding objects.
It adopts generative adversarial networks (GANs) to estimate realistic images.
\cite{wi2vi, multiview} estimate grayscale images of human motion and output video sequences.
They have an encoder-decoder architecture composed of a CSI encoder, feature translator, and image decoder.
These works adopt the framework of multimodal learning and try to establish an end-to-end image estimation pipeline from CSI.
However, while this framework is useful for estimating grayscale or RGB textures and illumination, we argue that these attributes are only loosely related to transmission links, making them difficult to reconstruct from CSI. 
Hence, we propose Wi-Depth.


\section{Proposed Wi-Depth Method}
We propose Wi-Depth, a deep learning model that generates depth images from CSI.
In this section, we introduce the proposed method designed based on the aforementioned solutions.

\subsection{Overview}
The overview of the proposed method is shown in Figure \ref{fig:overview}.
In order to learn the mapping from CSI to depth images, we adopt a two-step solution, involving a teacher-student architecture.
In the first step, the teacher learns the latent space of depth images.
In the second step, the student learns to map CSI into the same latent space as the teacher.
A mapping from CSI to the latent space is established by this means.

In the teacher training, the teacher network takes depth images as input, compresses the images into latent representations, and reconstructs the original depth images from the latent representations via the Variational Autoencoder (VAE) structure.
The teacher also outputs the shape, position, and depth of the moving objects via auxiliary decoders, facilitating the prediction of consistent components of the depth images.

In the student training, the teacher network is frozen and utilized.  Paired CSI and depth images are passed through the student and teacher simultaneously.
An encoder of the student is trained to output feature maps identical to those from the teacher in the teacher-student training scheme. 
In the end, the student estimates a depth image as well as the shape, depth, and position via the teacher's decoders.



The final model is composed of the encoders of the student and the decoders of the teacher.
Upon deployment, the model takes CSI as input and outputs a depth image in an end-to-end manner.
The auxiliary decoders for shape, position, and depth are also used to reinforce the reconstruction task.

\subsection{Preliminaries}

This study assumes an indoor environment where a Wi-Fi transmitter (AP) and receiver (computers or smartphones) are placed on the side of the sensing zone.
For example, the sensing zone covers the entrance of the room to capture a person entering or exiting. This motion is detected through changes in Wi-Fi transmission. 
We set one transmitter with 3 antennas and one receiver with 3 antennas, i.e., $N_{Tx}=3$, $N_{Rx}=3$.
The number of subcarriers in OFDM is $N_{sub}=30$.
The input of the proposed method is a window of CSI time series of length $N_{packet}=300$.
Since we preprocess CSI at the cost of eliminating $N_{Tx}$, the input CSI shape is $N_{packet} \times N_{sub} \times N_{Rx}$.

As input for training the proposed method, depth images are collected by a depth camera.



\subsection{Preprocessing}

\begin{figure}[t]
 \begin{center}
\vspace{0.2cm}
\includegraphics[scale=.7]{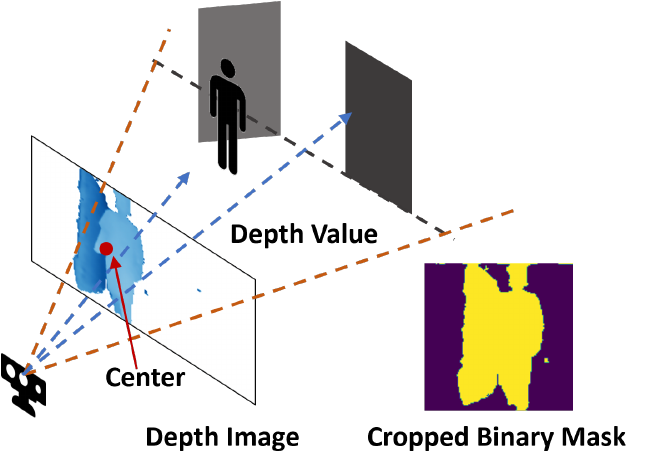}
\caption{Depth image components}
\label{fig:components}
 \end{center}
\end{figure}

Here we preprocess CSI and depth images used in the teacher-student learning. 
From depth images, we extract depth images of moving objects. 
In addition, we extract three core components used in the auxiliary tasks from the depth images of moving objects: the shape, position, and depth. 
Note that the shape refers to a cropped binary mask of the target, the position is the center coordinates indicating the azimuth of the target, and the depth is the average depth value representing  the distance of the target.
Figure \ref{fig:components} shows the core components for an example depth image.

The detailed preprocessing procedures are described in the experiment section. 

\subsection{Teacher-student Learning}

Wi-Depth is mainly composed of our proposed teacher-student learning architecture, as illustrated in Figure \ref{fig:structure}. 
The features of the architecture are: (i) image generation by leveraging a VAE, (ii) knowledge distillation in the latent space, and (iii) introducing auxiliary tasks for reconstructing the core components of depth images. 

\subsubsection{Image Generation with VAE}
Because the final output of our method is depth images, an image generator is required in our method.
This requires an image decoder whose input is a random variable (usually known as a latent representation) and the output is an image.
A traditional autoencoder (AE) is composed of an encoder and a decoder.
However, in the traditional AE, the encoder maps input data directly to a point in the latent space, and the decoder reconstructs the input from that point.
The latent space is not constrained to follow any specific distribution, which often results in a disorganized or sparse latent space.
As a result, interpolation between points in the latent space may lead to abrupt changes or invalid samples, as there is no guarantee that intermediate points correspond to valid or meaningful data.

In order to generalize to unseen images, we adopt a deep generative model, namely VAE\cite{vae}.
Compared to non-generative methods, the VAE adapts to a regularized latent space, enabling smooth interpolation of data points.
The latent space of a VAE is structured to follow a known probability distribution, typically a multivariate Gaussian.
During training, the encoder maps input data to a distribution in this latent space rather than a single point.
The decoder then reconstructs data by sampling from this distribution.
This encourages the latent space to be continuous and smooth, where nearby points correspond to similar data in the input space.
This structure allows for smooth interpolation between points in the latent space, resulting in a smooth transition between generated samples.

In addition, compared to other generative methods like generative adversarial networks (GANs) and diffusion models, a VAE is lightweight and easier to train.
For these reasons, we train a VAE with depth images to achieve a smooth latent space of object movements.

\subsubsection{Knowledge Distillation via Latent Space Mapping}
To bridge the gap between CSI and depth images, we perform knowledge distillation by utilizing the latent space.
The learned latent space from a VAE can be represented by two vectors, $\mu$ and $\sigma$,  respectively.
This latent space represents depth images, and presumably also the representation of object movements.
Regarding this, we map CSI features, which also represent object movements, to this latent space.

The connection between CSI and depth images is established by passing the output of CSI through a CNN-LSTM and then through a fully connected network to estimate $\mu$ and $\sigma$, which are generated by the VAE, using paired CSI matrices and depth images.
This approach solves the ``curse of dimensionality'' when one attempts to learn the direct mapping from high-dimensional CSI matrices to depth images.
The outputs of the CNN have much fewer dimensions than CSI, which are further condensed  by the LSTM.
Hence, the mapping from CNN-LSTM features to $\mu$ and $\sigma$ is much easier to learn.

\subsubsection{Introducing Auxiliary Tasks for Core Components of Depth Images}
Although a VAE learns latent representations of depth images, it is not easy to consistently capture the three core components of depth images: shape, depth, and position. Since we adopt a CNN-based VAE, the CNN is more sensitive to patterns of pixel value changes, but less sensitive to absolute values and pixel positions. 
As a result, the VAE may learn a good representation of shape in depth images, but a poorer representation of depth values and positions.

To improve the consistency of these core components, we introduce auxiliary tasks to guide the representation learning of shape, depth, and position. We attach multiple auxiliary decoders to the latent representation so that one latent vector is decoded simultaneously into a full image, a binary mask, the average depth value, and the center coordinates of the object. These auxiliary tasks aim to learn the core components, resulting in a latent space where each latent point represents a depth image and its consistent core components.

\subsection{Teacher Network}

\begin{figure}[t]
 \begin{center}
\hspace{-0.2cm}\includegraphics[scale=.4]{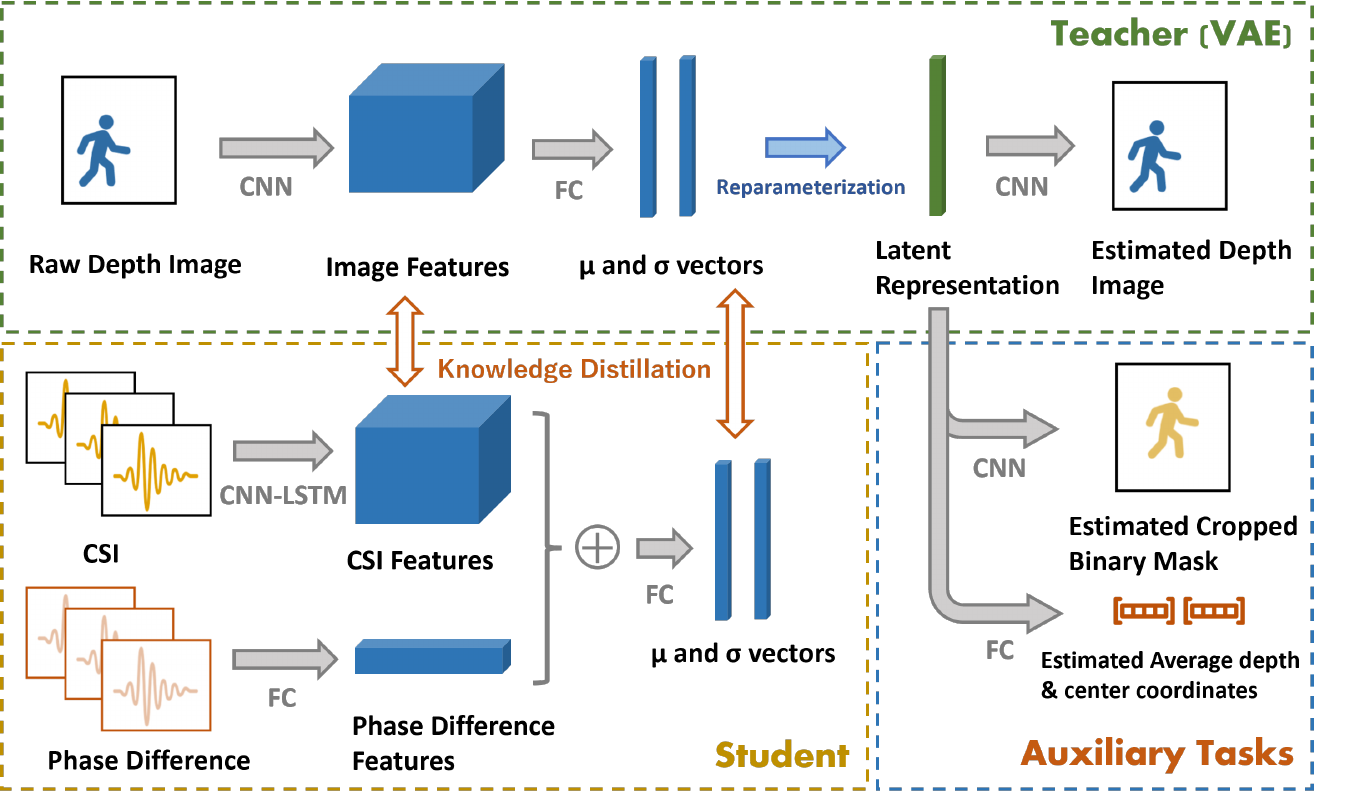}
\caption{Overview of teacher-student learning. All CNNs include 5 convolutional layers. The LSTM includes 2 layers. All FCs include 2-3 layers. Latent representations are generated from $\mu$ and $\sigma$ by reparameterization trick \cite{vae}. CSI features and phase difference features are flattened and concatenated. Average depth and center coordinates are estimated in one vector.}
\label{fig:structure}
 \end{center}
\end{figure}

As mentioned above, we adopt a teacher-student architecture to bridge the gap between CSI and depth images. 
As the foundation of the proposed model, we train a teacher network, depicted in Figure \ref{fig:structure}, to learn the latent space of depth images using a VAE. Unlike simple autoencoders that have no regularization on latent vectors, a VAE employs a distribution loss to regularize the learned latent distribution into a normal distribution. Since the learned latent distribution is constrained to a tight vicinity, it is possible to sample from this normal distribution and generate new samples, meaning that unseen images are also distributed within this vicinity. We leverage this characteristic to build our teacher network. 
The standard loss function of a VAE is as follows:
\begin{equation}
\begin{aligned}
\mathcal{L}_{VAE} &= \beta D_{KL}(P_{\hat{v}}\parallel Q_{u}) + \mathcal{L}_{recon}\\
&= \beta \sum_{\hat{v} }^{} P(\hat{v})\log(\frac{P(\hat{v})}{Q(u)} ) + \frac{1}{N} \sum_{i=1}^{N}(y_i-\hat{y_i})^2 
\label{eq:vae}
\end{aligned}
\end{equation}
where $D_{KL}(P_{\hat{v}}\parallel Q_{u})$ is the Kullback–Leibler (KL) divergence of the learned distribution against the normal distribution, $P_{\hat{v}}$ refers to teacher's learned latent distribution, $Q_{u}$ is the normal distribution,
$\mathcal{L}_{recon}$ refers to the reconstruction loss of the given input,
$\beta$ is the weight of the distribution loss,
$N$ is the batch size, and $y_i$ is the ground truth depth image, $\hat{y_i}$ is the estimated depth image.
As found by \cite{bvae}, when $\beta=0$, the VAE degrades to autoencoder.
The larger $\beta$ is, the more regularized the latent space becomes; meanwhile, the reconstruction quality deteriorates. 
We adopt 0.5 as a reasonable value for $\beta$.

In addition to the KL loss and reconstruction loss of a vanilla VAE, we add more terms to the loss function to aid in training, i.e., reconstruction losses from cropped binary masks, estimated positions, and estimated center coordinates. The extended VAE loss is as follows:

\begin{equation}
\begin{aligned}
\mathcal{L}_{VAEex} &= \beta \mathcal{L}_{KL} + w_1\mathcal{L}_{rimg} \\
&+ w_2\mathcal{L}_{cimg} + w_3\mathcal{L}_{depth} + w_4\mathcal{L}_{center} \\
&= \beta \sum_{\hat{v} }^{} P(\hat{v})\log(\frac{P(\hat{v})}{Q(u)} ) + \frac{w_1}{N} \sum_{i=1}^{N}(y_i-\hat{y_i})^2 \\
&+ \frac{w_2}{N} \sum_{i=1}^{N}-(\hat{y_i}\log(y_{ci})+(1-\hat{y_i}\log(1-y_{ci})))\\
&+\frac{w_3}{N} \sum_{i=1}^{N}(d_i-\hat{d_i})^2 +\frac{w_4}{N} \sum_{i=1}^{N}(c_i-\hat{c_i})^2 \\
\end{aligned}
\end{equation}
where $\mathcal{L}_{rimg}$, $\mathcal{L}_{cimg}$, $\mathcal{L}_{depth}$, and $\mathcal{L}_{center}$ are the reconstruction losses of depth images, cropped binary masks, average depth, and center coordinates, respectively.  
$w_1$ to $w_4$ are weights of corresponding loss terms,
$y_{ci}$ and $\hat{y_{ci}}$ are ground truth and estimated cropped binary masks,
and $d_i$, $\hat{d_i}$, $c_i$, $\hat{c_i}$ are ground truths and estimates of average depth value and center coordinates, respectively.
We adopt Mean-Squared-Error (MSE) loss as $\mathcal{L}_{rimg}$, $\mathcal{L}_{depth}$, and $\mathcal{L}_{center}$.
Binary Cross-Entropy (BCE) loss is adopted as $\mathcal{L}_{cimg}$.

By training with this loss function, the depth image encoder and the depth image decoder are updated together with the cropped image decoder and the center-depth decoder. Due to the regularization from the auxiliary decoders, the teacher learns a latent distribution of depth images with consistent shape, position, and depth. 
We adopt $\mathcal{L}_{VAEex}$ as the teacher loss $\mathcal{L}_{teacher}$.

\subsection{Student Network}
As mentioned above, the teacher is a generative model that learns the latent space of depth images. The student network, depicted in Figure \ref{fig:structure}, learns to map CSI into the latent space. To estimate depth images for moving objects, features related to motion are extracted from CSI. Conventional methods extract low-level contexts such as AoA, ToF, and DFS via Multiple Signal Classification (MUSIC) and Fourier transform. Such features contribute to developing tracking and action recognition methods. However, higher-level contexts related to movements are required to achieve our goal.

As OFDM provides redundancy in CSI\cite{redundancy}, we implement convolutional neural networks (CNN) \cite{cnn} and long-short term memory (LSTM) \cite{lstm} to extract spatial and temporal features from CSI.
Given input CSI in real and imaginary parts, CNN can be trained to cover the low-level contexts, as it is able to calculate differences of adjacent elements and select the maximum, similar to how MUSIC operates, though CNN is not strictly designed to mimic MUSIC. Furthermore, with deeper layers, CNN can extract higher-level contexts through a weighted sum and non-linear selection of the low-level contexts. The output of the CNN contains both low-level and high-level contexts. We then pass these features through an LSTM in time order, which can estimate movements based on how features vary in each time window.

This is achieved by an encoder with two branches that take CSI and phase difference as inputs. The CSI branch is composed of a CNN and an LSTM to extract spatial and temporal features from CSI. The phase difference branch is a simple fully connected network because the feature of phase difference is more explicit and does not require deep networks. The features of CSI and phase difference are combined in a late fusion manner and transformed to match the shape of the teacher's feature via fully connected layers.

To make the student learn the same distribution as the teacher, we adopt a three-level loss as follows:

\begin{equation}
\mathcal{L}_{student} = w_5\mathcal{L}_{feature} + w_6\mathcal{L}_{latent} + w_7\mathcal{L}_{gt}
\end{equation}
where $\mathcal{L}_{feature}$ is feature-level loss, $\mathcal{L}_{latent}$ is latent-level loss, $\mathcal{L}_{gt}$ is ground truth-level loss, and $w_5$ to $w_7$ are weights of the corresponding loss terms.

$\mathcal{L}_{feature}$ is calculated by:
\begin{equation}
\begin{aligned}
\mathcal{L}_{feature} &=\frac{1}{N} \sum_{i=1}^{N}(f_{ti}-f_{si})^2\\
&= \frac{1}{N} \sum_{i=1}^{N}[P_{TF}(y_i) - P_{SF}(x_i, \phi_i)]^2
\end{aligned}
\end{equation}
where $f_{ti}$ and $f_{si}$ are the features of the teacher and student, respectively.
$y_i$ is depth image, $x_i$ is CSI, $\phi_i$ is phase difference, $P_{TF}$ is the feature encoder of the teacher (CNN), and $P_{SF}$ is the feature encoder of the student (CNN-LSTM and FC).

$\mathcal{L}_{latent}$ is calculated by:
\begin{equation}
\begin{aligned}
\mathcal{L}_{latent} &=\frac{1}{N} \sum_{i=1}^{N}(z_{ti}-z_{si})^2\\
&= \frac{1}{N} \sum_{i=1}^{N}[R_{TZ}(f_{ti}) - R_{SZ}(f_{si})]^2\\
&\doteq \frac{1}{N}[\alpha\sum_{i=1}^{N}(\mu_{ti} - \mu_{si})^2 + (1 - \alpha) \sum_{i=1}^{N}(\sigma_{ti} - \sigma_{si})^2]
\end{aligned}
\end{equation}
where $z_{ti}$ and $z_{si}$ are the latent representations by the teacher and student respectively. 
$R_{TZ}$ and $R_{SZ}$ are the latent vector decoders (FC) of the teacher and the student. 
$\mu_{ti}$, $\sigma_{ti}$, $\mu_{si}$, and $\sigma_{si}$ are $\mu$ and $\sigma$ vectors estimated by the teacher and student.
$\alpha$ is the weight to balance $\mu$ and $\sigma$ losses.
We adopt the MSE for $\mathcal{L}_{feature}$ and $\mathcal{L}_{latent}$.

The feature-level loss is used to train an adaptive transformation of CSI-related features into depth image-related features. The latent-level loss forces the student to learn the same latent distribution as the teacher. The ground truth-level loss further guides the student with core components as follows:

\begin{equation}
\begin{aligned}
\mathcal{L}_{gt} &= w_2'\mathcal{L}_{cimg} + w_3'\mathcal{L}_{depth} + w_4'\mathcal{L}_{center}\\
&= \frac{w_2'}{N} \sum_{i=1}^{N}-(\hat{y_{si}}\log(y_{ci})+(1-\hat{y_{si}}\log(1-y_{ci})))\\
&+\frac{w_3'}{N} \sum_{i=1}^{N}(d_i-\hat{d_{si}})^2 +\frac{w_4'}{N} \sum_{i=1}^{N}(c_i-\hat{c_{si}})^2 \\
\end{aligned}
\end{equation}
where $\hat{y_{si}}$, $\hat{d_{si}}$, $\hat{c_{si}}$ are cropped binary masks, average depth value, and center coordinates generated from student's latent estimates, respectively.
As with the teacher, we adopt MSE for $\mathcal{L}_{depth}$ and $\mathcal{L}_{center}$, BCE for $\mathcal{L}_{cimg}$.

Note that all decoders from the teacher's side are frozen during the training of the student, so the ground truth-level losses only contribute to updating the student’s encoder.

To train the teacher-student architecture, we first train the teacher network. The student network is trained afterward. Paired CSI and depth images are used in student training. A depth image is passed through the teacher to obtain the teacher's latent estimate. CSI and extracted phase difference are passed through the student to obtain the student's latent estimate. Cropped binary images, average depth values, and center coordinates are used to compute the ground truth-level loss.

\begin{figure}[ht]
 \begin{center}
\includegraphics[width=0.80\linewidth]{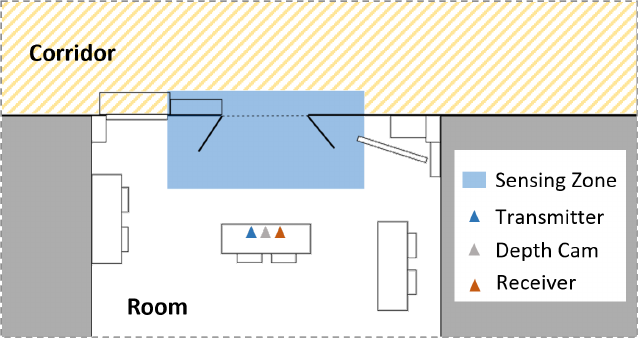}
\caption{An example of experimental setting. Experiments were conducted at four rooms of different shapes and with different furniture settings.}
\label{fig:setting}
 \end{center}
\end{figure}

\section{Experiment and Evaluation}

\subsection{Experimental Environments and Data Collection Procedures}
We conducted data collection experiments in four real environments. An example of the experimental setup is shown in Figure \ref{fig:setting}. 
We set the transceiver at a desktop near the entrance of the room to work as surveillance that captures the depth images of human entering and exiting the room.
The sensing zone is 2 m $\times$ 4 m.
The depth camera used for collecting ground truth data was positioned on the centerline, 3 m away from the farthest side of the sensing zone. The entrance is 2.5 m from the camera. Given the camera's viewing angle and valid sensing range, this setup provides a natural view of moving humans in practical use.
Next to the camera, the antennas of the Wi-Fi AP and receiver were positioned 30 cm apart. The desktop holding all sensors is 75–80 cm high.

When collecting data, the Wi-Fi transmitter sent packets at 1 kHz. The transmission was established on a 5.32 GHz channel with a bandwidth of 20 MHz. Both the transmitter and the receiver had three antennas. An OFDM link with 30 subcarriers was extracted using an Intel 5300 NIC.

The clocks of the Wi-Fi receiver and the depth camera were calibrated with an error margin of 10 ms. The depth camera recorded ground truth images at a resolution of 848 × 480 pixels and 30 fps. Ideally, one depth image would correspond to 33 CSI packets. However, in practice, we find that the fps of the depth camera is quite unstable, causing frame drops. This required longer time windows so that the missing motion could still be aggregated in neighboring samples. Since the average walking speed of an adult human is 1.2 m/s, we selected 300 packets (0.3 sec) as the window length for CSI. Within 0.3 seconds, we were able to observe motion while preventing it from becoming too complex. In line with causality, we aligned depth images at the end of the nearest CSI windows by timestamp matching.

We collected datasets with 6 volunteers (4 male and 2 female) in four environments with different areas and furniture. Each volunteer was asked to enter and exit the test room, both with and without opening the door. They perform actions in their own accustomed ways. A total of 203,428 samples (CSI and paired depth images) were collected.

\subsection{Preprocessing}
We performed data preprocessing on CSI and depth images to extract dynamic components and mitigate the impact of environmental factors.

\subsubsection{CSI}
In practice, CSI is contaminated by several offsets. Specifically, Sampling Frequency Offset (SFO), Carrier Frequency Offset (CFO), and so forth \cite{offset}. These offsets cause fluctuations in the CSI phase. Additionally, the static component of CSI overwhelms the dynamic component in amplitude, making it harder to track changes using CSI.

To address these issues, we apply a three-step preprocessing method on CSI:
1) Divide the CSI matrix by the values of a specific transmitter antenna (e.g., $Tx_0$) to remove the receiver-side phase offsets.
Because all of the $N_{Tx} \times N_{Rx}$ links share the same receiver-side offsets, the division can remove the offsets despite the loss of $N_{Tx}$ information.
The preprocessed $N_{packet} \times N_{sub} \times N_{Rx}$ CSI matrix still contains necessary information because the dimensions of multiple $N_{Rx}$ and $N_{sub}$ are preserved.
2) Apply a high-pass filter to the CSI to extract dynamic components.
3) Apply a Savitzky–Golay filter to each time window of the CSI to remove high-frequency noise and improve robustness against noise.

In addition, we calculate phase differences from the CSI.
By performing SVD on the complex CSI matrix, reshaped into $N_{Rx} \times (N_{sub} \times N_{packet})$, and calculating the conjugate multiplication of neighboring eigenvectors, we obtain the major phase differences among the receiver antennas. 
Similarly, by performing SVD on the CSI, reshaped into $N_{sub} \times (N_{Rx} \times N_{packet})$, and calculating the conjugate multiplication, we obtain the major phase differences among the subcarriers. These phase differences are closely related to AoA and ToF, and are therefore important in estimating the position and depth of the subject.
To smooth the phase differences in each segment, a temporal median filter is applied to remove noise. 


\subsubsection{Depth Image}
In terms of depth image preprocessing, there are four main steps:
1) Apply general filters, such as hole-filling.
2) Remove depth values that exceed the maximum sensing range of the depth camera. 
3) Apply a median filter along the time axis to remove flicker noise. Moreover, the temporal median filter can extract the active part of the image, such as the silhouette of humans and objects in motion.
4) Since the pixel values of depth images represent depth in millimeters, we apply a threshold and normalize the values to $[0, 1]$ for training purposes.

\textcolor{black}{
The ground truths of the core components for auxiliary tasks, i.e., the shape, center, and depth, are extracted from the depth images as follows. Fundamentally, we focus on the moving object in the depth image by generating a bounding box around the object using computer vision methods. This is done by calculating the contours of connected areas in the image and identifying the one with the largest area. With the bounding box, the center point coordinates are calculated. Meanwhile, the average depth of the bounded area is computed. The cropped depth image is then binarized based on whether each pixel value is greater than 0. Finally, a 128 × 128-pixel binary mask of the moving object is cropped from the depth image according to the center point.
}

\subsection{Evaluation Methodology}

Because we have six subjects in each environment, we employed leave-one-subject-out cross-validation by environment. 
That is, one subject was used as the test subject and the remaining five subjects were used as training subjects.

We evaluated the proposed method by assessing the quality of the generated depth images. For quantitative metrics, we calculated the Soft IoU between the generated depth images and ground truth images as the primary criterion. Soft IoU is calculated based on $1 -  \frac{Intersection}{Union}$ on the depth images. By considering both pixel values and pixel positions, this criterion is particularly crucial for object detection and semantic segmentation tasks \cite{softiou}. It is equally vital for depth images because we place significant emphasis on depth values, as well as the precise shape and positional information.
The MSE between the generated depth images and ground truth images was also calculated for a simple evaluation metric. 
Additionally, we calculated errors for each core component: shape, position, and depth.
We applied 2D correlation to each estimated depth image against its ground truth to determine how much translation was required to achieve the maximum correlation. The Euclidean distance from the ground-truth center to the estimated center, normalized by image size, was calculated as the position error. 
After translating the estimated depth image to the point of maximum correlation, we  calculated the IoU of the binary masks between the matched images as the shape error.
As for the depth error, we calculated the MSE between histograms of estimated depth images and ground truth, as to consider the continuity of depth values regardless of shape and position.
\textcolor{black}{
Note that the frequency values of a histogram are normalized by dividing the total number of pixels. The number of bins is 50. 
} 


\subsection{Comparative Methods}
As this work is the first to estimate depth images from CSI, there is little existing work for comparison. However, we can adapt existing methods that estimate RGB or grayscale images from CSI to our task. For example, Wi2Vi, proposed by \cite{wi2vi} and \cite{multiview}, estimates grayscale image sequences from CSI.
\textcolor{black}{We reproduced the Wi2Vi structure and modified the number of channels in the input layer to fit our dataset.}

As baseline models, we employed end-to-end AE and VAE to perform the same task, meaning that we passed CSI through the model and obtained depth images as output during both training and validation. The loss function for the AE was MSE, and for the VAE, it was the standard VAE loss, as shown in Eq. \ref{eq:vae}.

To validate the structural components of the proposed method, we also designed baseline models using a teacher-student architecture, named TSAE and TSVAE. 
In this architecture, two types of teacher networks were used: AE and VAE. The learned latent space of the teacher was used to train the student, as in the proposed method. However, we applied the basic loss functions for the AE and VAE teachers. 

Considering training costs, we randomly selected 20\% of samples from the dataset for each training epoch. The selected samples remained identical across all proposed and comparative methods.
\textcolor{black}{For validation, all the test samples were used.}


\subsection{Results}


\begin{table}[t]
\caption{Quantitative metrics of the methods}
\centering
\label{tab:comparative}
\scalebox{1.0}{\begin{tabular}{l|lllll}
                  &  Image error           &   Image error     & Depth    & Shape & Position  \\ 
                  & (MSE)            & (Soft IoU)       & error (\%)    & error & error \\ \hline
\textbf{Wi-Depth} & \textbf{0.022} & \textbf{0.836} & \textbf{0.012} & \textbf{0.607}      & 0.170          \\
AE               & --          & --          & --          & --               & --         \\
VAE               & 0.032          & 0.880          & 0.134          & 0.661               & \textbf{0.156}          \\
Wi2Vi             & 0.164          & 0.939          & 2.143          & 0.902               & 0.176          \\
TSAE              & 0.064          & 0.949          & 1.806          & 0.894               & 0.165         \\
TSVAE             & 0.025          & 0.889          & 0.040          & 0.721               & 0.179        
\end{tabular}}
\end{table}

\begin{figure}[t]
 \begin{center}
\includegraphics[width=0.8\linewidth]{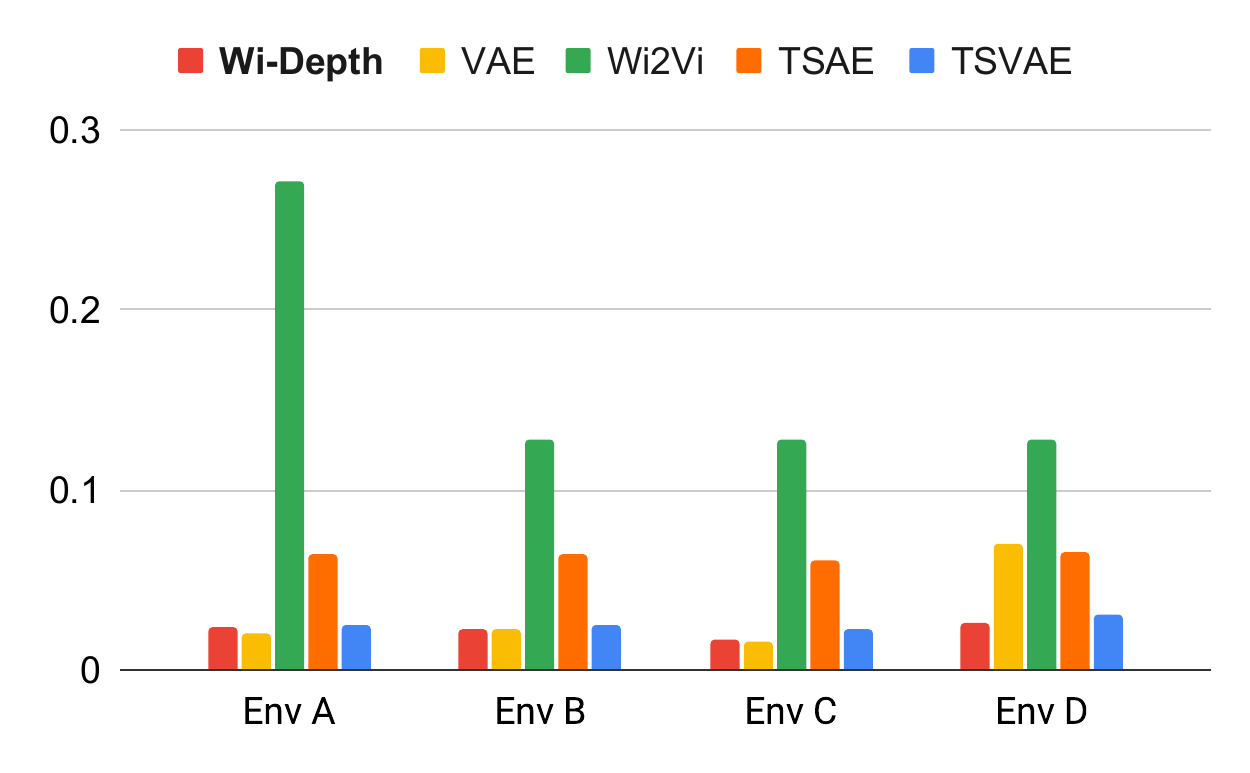}
\caption{Average MSE per environment}
\label{fig:bar_mse}
 \end{center}
\end{figure}

\begin{figure}[t]
 \begin{center}
\hspace{1cm}\includegraphics[width=0.8\linewidth]{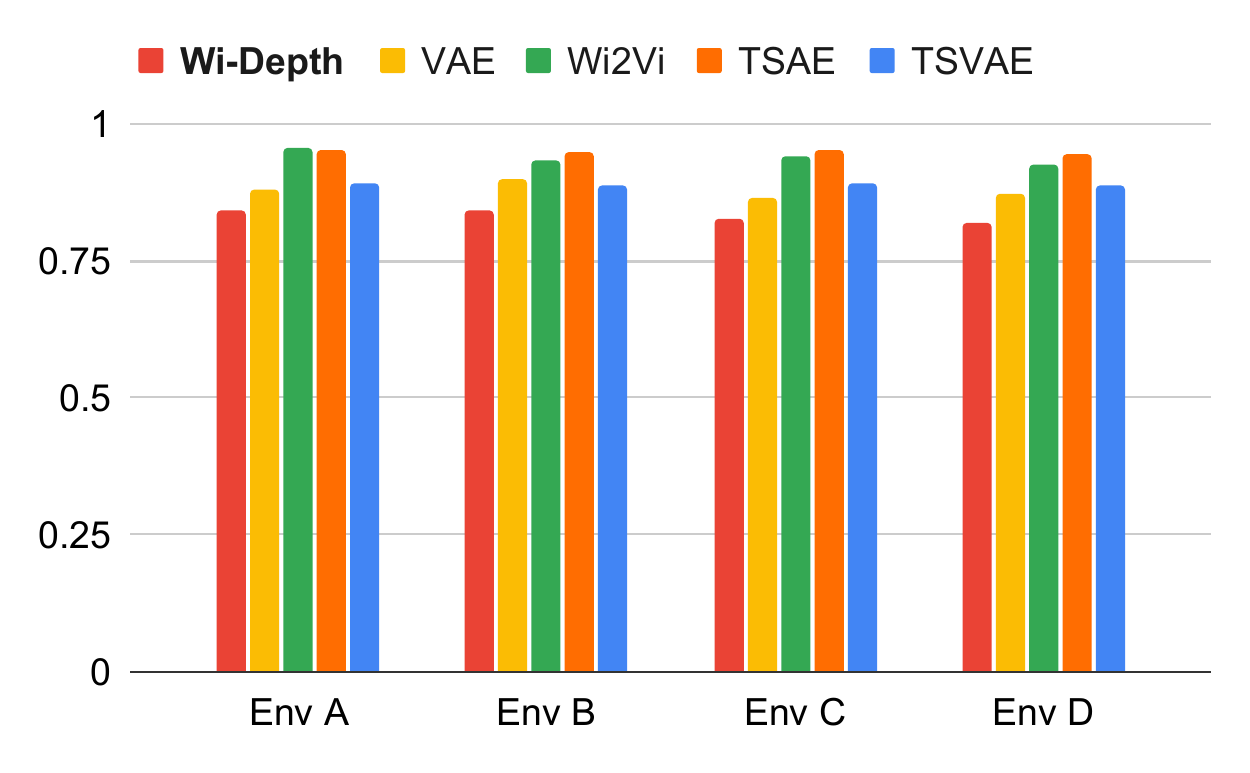}
\caption{Average Soft IoU per environment}
\label{fig:bar_softiou}
 \end{center}
\end{figure}

Table \ref{tab:comparative} shows the overall comparison of the depth images generated by the proposed Wi-Depth method and comparative methods. The results for each environment were averaged across all subjects. 
Fig. \ref{fig:bar_mse} and Fig. \ref{fig:bar_softiou} provide a detailed comparison of MSE and Soft IoU averaged across test subjects in each environment.

\begin{figure}[t]
 \begin{center}
    \hspace{-0.6cm}\includegraphics[width=1.06\linewidth]{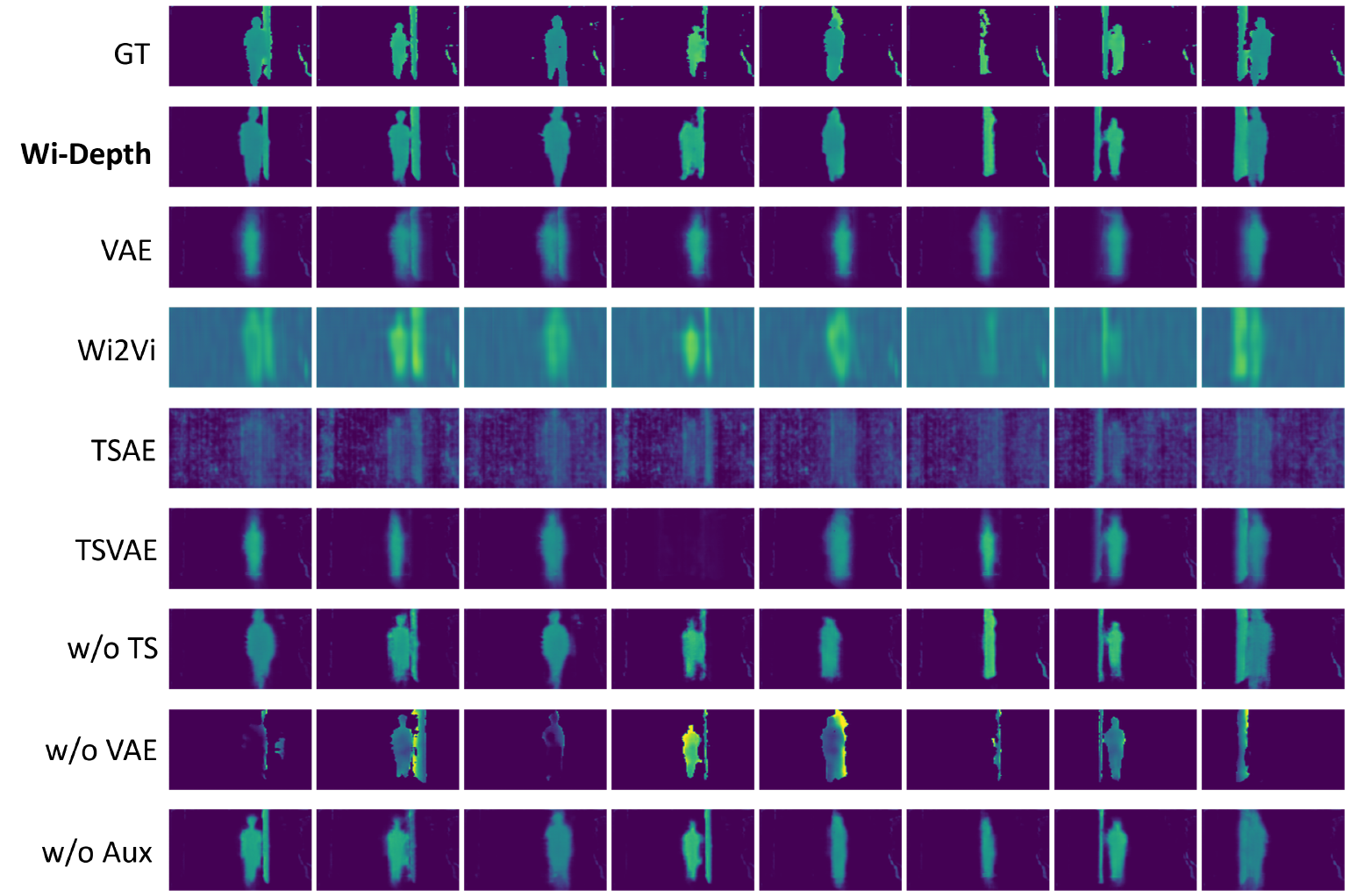}

\caption{Example of generated depth images}
\label{fig:res_visual}
 \end{center}
\end{figure}

The proposed method achieved the lowest Soft IoU and MSE on average, demonstrating the highest estimation performance among all compared methods in the quantitative analysis. 
However, for depth image reconstruction, providing recognizable visual quality is more important. The visual quality of the generated depth images is shown in Fig. \ref{fig:res_visual}.
Wi2Vi generated blurry images, often with a conspicuous non-zero background. The baseline VAE also produced blurry images, with the door shape in the context notably absent. 
\textcolor{black}{
The VAE method generated the same human shape in several different images, suggesting that this method learns to generate an averaged, blurred shape to achieve small reconstruction errors. 
In fact, this method achieved low MSEs as shown in Table \ref{tab:comparative} and Fig. \ref{fig:bar_mse}, indicating the limitation of the simple MSE metric. 
}
\textcolor{black}{Because the baseline AE was practically unable to train (yielding blank images or NaN values after training), indicating that the straightforward mapping from CSI to depth images was actually difficult to learn, we could not calculate the evaluation metrics for AE.}
The failure of these three models suggests that generating depth images from CSI in an end-to-end manner is challenging. During training, the decoder is optimized to generate images, but the gradient must be back-propagated to the input layer, which handles a completely different modality (i.e., CSI). This creates an optimization trade-off between the encoder and the decoder. Even though Wi2Vi incorporates a ``feature interpreter'' between the encoder and decoder \cite{wi2vi}, this issue was not mitigated.

Teacher-Student AE (TSAE) failed to estimate visually recognizable images. Although the AE teacher was trained to estimate correct depth images (not shown in the figures), the student was unable to learn the latent representation of the images, resulting in aimless image generation. This is because the latent representation learned by the AE teacher was not regularized, making it difficult for the student to extract meaningful latent representations over a large span. 
Teacher-Student VAE (TSVAE) was able to generate visually comparable images, but the consistency of the core components was poor. It failed to integrate the correct shape, depth, and position simultaneously. This approach can be considered a prototype of the proposed model. However, since neither the teacher nor the student was assisted by auxiliary tasks, the knowledge of the core components was not explicitly encoded in the latent representation, making it difficult to estimate depth images that align with the ground truth.
The shortcomings of the Teacher-Student AE and Teacher-Student VAE reveal the importance of learning the core components in a regularized latent representation.

In contrast, Wi-Depth demonstrates superior visual quality in the generation of depth images compared to the comparative methods. While other techniques tend to produce images with noticeable blurriness or structural inaccuracies, our approach generates sharper, more detailed images, accurately preserving structural geometry and ensuring consistency in depth and position. This improvement is evident both qualitatively and quantitatively, as our method reduces common artifacts found in the comparative approaches.

Moreover, Table \ref{tab:comparative} shows the quantitative results regarding core components of shape, depth and position.
As illustrated in this table, Wi-Depth outperforms the others in terms of shape error and depth error, suggesting that the model generates depth images by understanding core components of the scene, rather than relying solely on pixel-level generation, as seen in other techniques.
\textcolor{black}{
Although the VAE and TSAE methods outperformed Wi-Depth in the position error, the errors across the five methods were quite similar. 
In addition, Fig. \ref{fig:res_visual} shows that all the methods precisely predicted the object positions, indicating that the object coordinates are easily learned. 
}


Nonetheless, none of the comparative methods is able to consistently perform well across all criteria.
\textcolor{black}{This highlights the challenging nature of estimating consistent depth images from CSI, and underscores the effectiveness of Wi-Depth as a significant advancement in the field of CSI sensing.}




\subsection{Ablation Study}
We tested the structural validity of the proposed method through an ablation study. The teacher-student architecture, VAE teacher, and auxiliary tasks are three key components of the proposed model. 
Therefore, we trained and evaluated three variations of the proposed method, each with one of these key components removed. These variations are referred to as w/o TS (without Teacher-Student architecture), w/o VAE (without the VAE teacher, using AE instead), and w/o Aux (without auxiliary tasks). We used the same validation metrics as those applied in the comparative methods for the ablation study. 

\begin{table}[]
\caption{Quantitative metrics of ablation study in Environment A}
\centering
\label{tab:ablation}
\scalebox{1.0}{\begin{tabular}{l|lllll}
                  &  Image error           &   Image error     & Depth    & Shape & Position  \\ 
                  & (MSE)            & (Soft IoU)       & error (\%)    & error & error \\ \hline
\textbf{Wi-Depth} & 0.024 & \textbf{0.840} & \textbf{0.010} & \textbf{0.607} & \textbf{0.158} \\
w/o TS            & \textbf{0.023} & 0.844          & 0.015          & 0.647          & 0.161          \\
w/o VAE           & 0.026 & 0.880          & 0.012          & 0.703          & 0.217          \\
w/o Aux           & 0.026 & 0.877          & 0.017          & 0.721          & 0.191         
\end{tabular}}
\end{table}

The bottom part of Fig. \ref{fig:res_visual} shows the generated depth images.
Table \ref{tab:ablation} shows the quantitative results.
First, without the teacher-student architecture, the model degrades into an end-to-end VAE, assisted by auxiliary tasks. As shown in the figure, the model estimates slightly blurred images and suffers from a loss of consistency. However, compared to the baseline VAE, the inclusion of auxiliary tasks does improve the structural integrity of the estimated object shapes, particularly in retaining the door shape within the context.

Moreover, without the VAE-based teacher, the model degrades to a Teacher-Student AE, assisted by auxiliary tasks.
The errors increased across almost every metric, showing that the VAE is vital to effective space latent learning.
The model even suffered from a loss of continuity in depth values, evidenced by unexpected changes in depth. 
\textcolor{black}{The deterioration in Soft IoU, shape error, and position error, noticeably demonstrates the vitality of a regularized latent space.}

Finally, in the absence of the auxiliary tasks, the model degrades into Teacher-Student VAE with additional phase difference input. As mentioned earlier, without the knowledge of core components learned through auxiliary tasks, the consistency of the generated depth images deteriorates. This results in visual artifacts such as inaccurate or blurred shapes, erroneous depth estimates, and misaligned object positions. This is also reflected by the simultaneous significant increase in depth, shape, and position errors.

\textcolor{black}{In the ablation study, Wi-Depth not only achieved the best performance in terms of quantitative criteria but also produced the highest visual quality. This demonstrates the structural effectiveness of the combination of the three components.}





\subsection{Limitations \& Discussion}

\begin{figure}[t]
 \begin{center}
\includegraphics[width=1.0\linewidth]{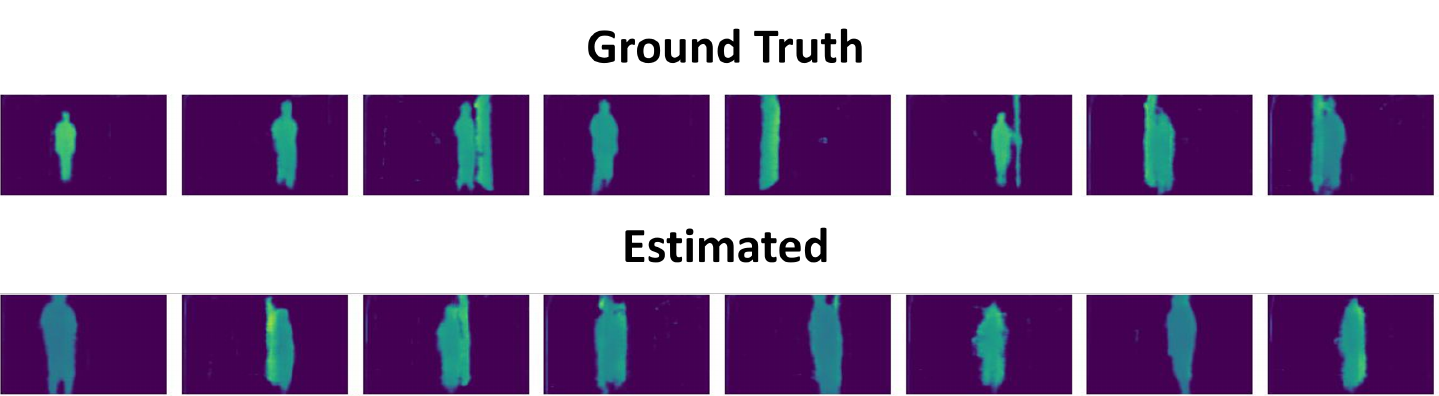}
\caption{Example of generated depth images by environment-independent model}
\label{fig:visual_env}
 \end{center}
\end{figure}

The experiment assumes environment-dependent models because Wi-Fi propagation heavily depends on environmental factors such as the positions of walls and Wi-Fi devices. Here, we investigate the performance of Wi-Depth when trained in an environment-independent manner. For example, Environment A was selected as a test environment, the Wi-Depth model was trained on data from the remaining environments. The mean Soft IoU of the environment-independent model for reconstructed images was 0.923, which was poorer than that of the environment-dependent model, which had a mean Soft IoU of 0.836.
\textcolor{black}{
Fig. \ref{fig:visual_env} shows an example image generated by the environment-independent model, indicating that the object positions were incorrectly estimated, although the shape and depth estimates were reasonable in many cases. This discrepancy may be caused by slight differences in the positional and directional relationships between the depth cameras and the Wi-Fi devices across different environments. Automatic calibration of these differences could be an important technique for achieving environment-independent depth image reconstruction. 
}

\section{Conclusion}
This work proposed Wi-Depth, the first approach to effectively estimate depth images from Wi-Fi CSI, via a teacher-student VAE with auxiliary tasks.
The performance of Wi-Depth was validated by leave-one-subject-out experiments in four real environments.
Compared to existing techniques, the proposed method excels in estimating depth images of decent visual quality with an low error, as well as high consistency in terms of core components of shape, depth, and position.

The structural effectiveness of the proposed method was further validated in the ablation study. The three key components--teacher-student architecture, VAE-based teacher, and auxiliary tasks focused on core components--proved to be indispensable and effective in generating visually recognizable depth images.


\bibliographystyle{ieeetr}

\bibliography{depth_imaging_ref}

\end{document}